\pdfoutput=1

\documentclass[11pt]{article}

\usepackage[review]{EMNLP2023}

\usepackage{times}
\usepackage{latexsym}

\usepackage[T1]{fontenc}

\usepackage[utf8]{inputenc}

\usepackage{microtype}

\usepackage{inconsolata}

\usepackage{url}
\usepackage{hyperref}
\usepackage{multicol}
\usepackage{multirow}
\usepackage{booktabs}
\usepackage{algorithm}
\usepackage{algorithmic}
\usepackage{graphicx}
\usepackage{amsmath}
\usepackage{xcolor}
\usepackage{listings}
\usepackage[utf8]{inputenc}
\definecolor{orange}{rgb}{1,0.5,0}
\definecolor{mdgreen}{rgb}{0.05,0.6,0.05}
\definecolor{mdblue}{rgb}{0,0,0.7}
\definecolor{dkblue}{rgb}{0,0,0.5}
\definecolor{dkgray}{rgb}{0.3,0.3,0.3}
\definecolor{slate}{rgb}{0.25,0.25,0.4}
\definecolor{gray}{rgb}{0.5,0.5,0.5}
\definecolor{ltgray}{rgb}{0.7,0.7,0.7}
\definecolor{purple}{rgb}{0.7,0,1.0}
\definecolor{lavender}{rgb}{0.65,0.55,1.0}

\definecolor{delim}{RGB}{20,105,176}
\definecolor{numb}{RGB}{106, 109, 32}
\definecolor{string}{rgb}{0.64,0.08,0.08}

\colorlet{punct}{red!60!black}
\definecolor{background}{HTML}{EEEEEE}
\definecolor{delim}{RGB}{20,105,176}
\colorlet{numb}{magenta!60!black}

\floatstyle{plain}
\newfloat{json}{thp}{lop}
\floatname{json}{JSON}
\lstdefinelanguage{json}{
    basicstyle=\small\ttfamily,
    showstringspaces=false,
    breaklines=true,
    frame=lines,
    backgroundcolor=\color{background},
    literate=
     *{0}{{{\color{numb}0}}}{1}
      {1}{{{\color{numb}1}}}{1}
      {2}{{{\color{numb}2}}}{1}
      {3}{{{\color{numb}3}}}{1}
      {4}{{{\color{numb}4}}}{1}
      {5}{{{\color{numb}5}}}{1}
      {6}{{{\color{numb}6}}}{1}
      {7}{{{\color{numb}7}}}{1}
      {8}{{{\color{numb}8}}}{1}
      {9}{{{\color{numb}9}}}{1}
      {:}{{{\color{punct}{:}}}}{1}
      {,}{{{\color{punct}{,}}}}{1}
      {\{}{{{\color{delim}{\{}}}}{1}
      {\}}{{{\color{delim}{\}}}}}{1}
      {[}{{{\color{delim}{[}}}}{1}
      {]}{{{\color{delim}{]}}}}{1},
}

\usepackage{amsmath,amsfonts,bm}

\def\eqref#1{equation~\ref{#1}}

\def\1{\bm{1}}

\DeclareMathAlphabet{\mathsfit}{\encodingdefault}{\sfdefault}{m}{sl}
\SetMathAlphabet{\mathsfit}{bold}{\encodingdefault}{\sfdefault}{bx}{n}

\newcommand{\softmax}{\mathrm{softmax}}

\title{Semi-Structured Object Sequence Encoders}

\author{\bf{Rudra Murthy V$^1$, Riyaz Bhat$^1$, Chulaka Gunasekara$^1$, Siva Sankalp Patel$^1$}\\
\bf{Hui Wan$^1$, Tejas Indulal Dhamecha$^2$, Danish Contractor$^1$, Marina Danilevsky$^1$} \\
\\
$^1$IBM Research AI \\
$^2$Microsoft India Development Center \\
rmurthyv@in.ibm.com,\{riyaz.bhat,chulaka.gunasekara,siva.sankalp.patel\}@ibm.com,
\\hwan@us.ibm.com, tdhamecha@microsoft.com, danish.contractor@ibm.com, mdanile@us.ibm.com}

\begin{document}
{\makeatletter\acl@finalcopytrue
  \maketitle
}
\nolinenumbers
\begin{abstract}
In this paper we explore the task of modeling semi-structured object sequences; in particular, we focus our attention on the problem of developing a structure-aware input representation for such sequences. Examples of such data include user activity on websites, machine logs, and many others. This type of data is often represented as a sequence of sets of key-value pairs over time and can present modeling challenges due to an ever-increasing sequence length. 
We propose a two-part approach, which first considers each key independently and encodes a representation of its values over time; we then self-attend over these value-aware key representations to accomplish a downstream task. This allows us to operate on longer object sequences than existing methods. We introduce  a novel shared-attention-head architecture between the two modules and present an innovative training schedule that interleaves the training of both modules with shared weights for some attention heads. Our experiments on multiple prediction tasks using real-world data demonstrate that our approach outperforms a unified network with hierarchical encoding, as well as other methods including a {\em record-centric} representation and a {\em flattened} representation of the sequence.
\end{abstract}

\section{Introduction}

\begin{figure*}
    \centering
    \includegraphics[scale=0.85]{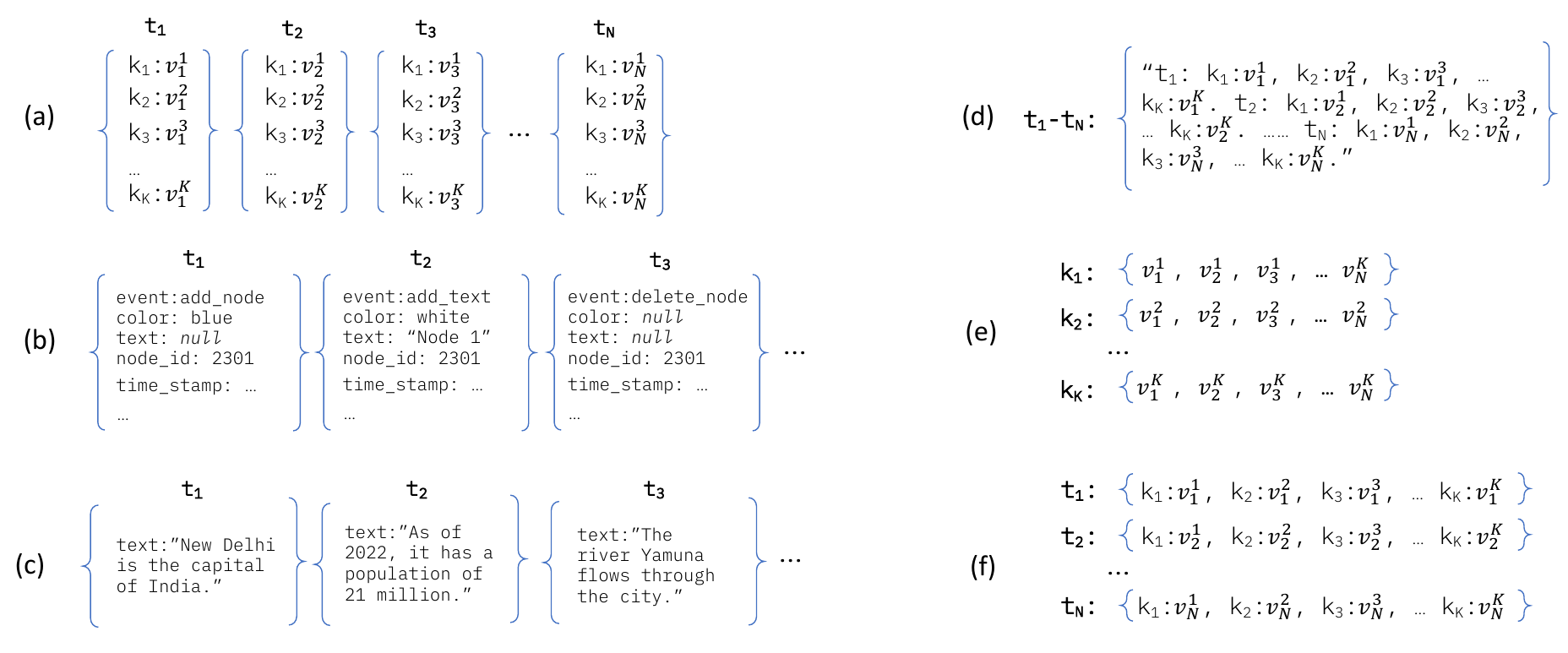}
    \caption{Semi-Structured Object Sequences: (a) Generic representation of a sequence of semi-structured objects, consisting of multiple key-value pairs at time steps $t_1 \dots t_N$. (b) Example object: a sequence of events triggered by the use of a graphical user interface. (c) Viewing a text paragraph as a sequence of sentences. (d) Encoding the example object in (a) by \textit{flattening}. (e) Encoding (a) by encoding a representation of the values for each key (to be followed by a key aggregation step, not shown). (f) Encoding (a) using a \textit{record-centric} representation for each time step.}
    \label{fig:samples}
\end{figure*}

Semi-structured object sequences comprise a significant portion of the myriad of data created daily. This data usually has a temporal aspect, with the data created sequentially and representing events happening in some order. More generally, the data is a sequence of structured objects, each represented by a set of key-value pairs that encode the attributes of the object 
(Figure \ref{fig:samples}(a)). Examples include recordings of user interactions with websites, logs of machine activity, shopping decisions made by consumers, and many more (Figure \ref{fig:samples}(b)). The data is usually stored in semi-structured formats such as JSONs, or tabular forms. 

In this paper, we explore the task of modeling semi-structured object sequences; in particular, we focus our attention on the problem of developing a structure-aware input representation for such sequences. 
If we think of the parallel to natural language data, we would treat each sentence of a text (Figure \ref{fig:samples}(c))
akin to the set of key-value pairs at a particular time step. 

\paragraph{The challenge of sequence length:}
A trivial method of representing such sequences would be to {\em flatten} each structured object and view its constituents as individual {\em words} for tokenization in natural language (Figure \ref{fig:samples} (d)) \footnote{with markers to indicate boundaries for each structured object}. However, this causes the sequence length to become extremely large (thousands of tokens) when operating on real-word semi-structured sequences. For instance, in our study of semi-structured objects from user-interaction sessions on software from a large cloud-based service provider, we found these objects could contain 11 fields in average. The values of these fields include timestamps, identifiers, log messages, etc., with an average of 5 words each. A session length of $15$ minutes results in $105$ such session objects, amounting to nearly $5,775$ words which would further increase the sequence length after sub-words are created. Thus, it quickly becomes clear that one of the main challenges of modeling the data is the sequence length;  each semi-structured object in a sequence contains many keys and values, and each sequence contains many such objects.  

\paragraph{Key-centric representation:} To address these challenges we use a modular, two-part hierarchical encoding strategy. 
First, we decompose the sequence of semi-structured objects into independent sets of sequences based on  keys. This allows us to consider each key separately (Figure \ref{fig:samples}(e)), and encode a representation of how the values of that key evolve over time (we refer to this as Temporal Value Modeling -- TVM). This may be achieved using any encoder.\footnote{We use BERT and Longformer in our experiments} We then self-attend over the set of the key encodings to create a representation of the entire structured object sequence (referred to as Key Aggregation -- KA). 
 
\paragraph{Advantages:} This key-centric  perspective of encoding semi-structured sequences has many advantages as compared to {\em flattening} and {\em record-centric representations} (Figure \ref{fig:samples}(f)). Decoupling the keys allows us to support an arbitrary number of keys\footnote{Real-world data can have hundreds of keys in each object.} since each key-sequence is encoded independently. So, the key-representations created during Temporal Value Modeling can support longer sequences than what would have been impossible with flattening (due to memory constraints). Moreover, this encoding strategy also 
 accommodates input sequences that may be considered non-structured -- e.g, natural language text as sequences of words in sentences (Figure \ref{fig:samples}(c)). Specifically, if we consider a sequence of structured objects where each structured object consists of only one key, whose values contain a {\em sentence}, then our TVM effectively encodes the text sequence using whatever encoder has been employed.\footnote{The key aggregation step, in this case, is redundant.} 

The decoupling of keys and the use of two independent encoders - TVM for value-aware key representations, and KA for aggregating key-representations for a downstream task - requires that information be shared between the two networks. So that the key-representations generated by TVM can be optimized for downstream tasks via the KA. 
To facilitate this, we share a few sets of attention heads between the two networks. First, we pre-train the TVM network with shared attention heads in place. We then use the frozen representations from this network to initialize the KA network, which has its own untrained attention heads, and the shared attention heads from the TVM network as part of its trainable parameters. We utilize a training schedule that interleaves the training of both modules to iteratively train them. Doing so allows the TVM and KA modules to create richer representations of keys, informed by their importance, for the downstream task. We find that this novel iterative two part-training results in better performance compared to a unified network with hierarchical encoding (with no attention-head sharing) as well as other methods, that either use a {\em flattened} representation or a {\em record-centric} representation of the sequence \cite{velivckovic2018graph,Mizrachi2019CombiningCF,de2021transformers4rec}. 




\noindent {\bf Contributions:} Our work addresses the challenges of encoding semi-structured object sequences: (i) we propose a two-part approach that separately encodes the evolution of the values for each key, followed by aggregation over key-representations to accomplish downstream  tasks; (ii) we present a novel approach for sharing attention heads between the components; (iii) we compare our approach against baselines such as sequence flattening, joint encoding, and record-centric sequence representations; and (iv) we present detailed experiments on several datasets and tasks to demonstrate the advantages of our approach. To the best of our knowledge this is the first work that develops a  framework that allows training models for tasks using long and large semi-structured object sequences. 

\section{Modeling}

We now describe our approach for modeling key-value semi-structured object sequences. 
\begin{figure*}
    \centering
    \includegraphics[width=0.7\linewidth]{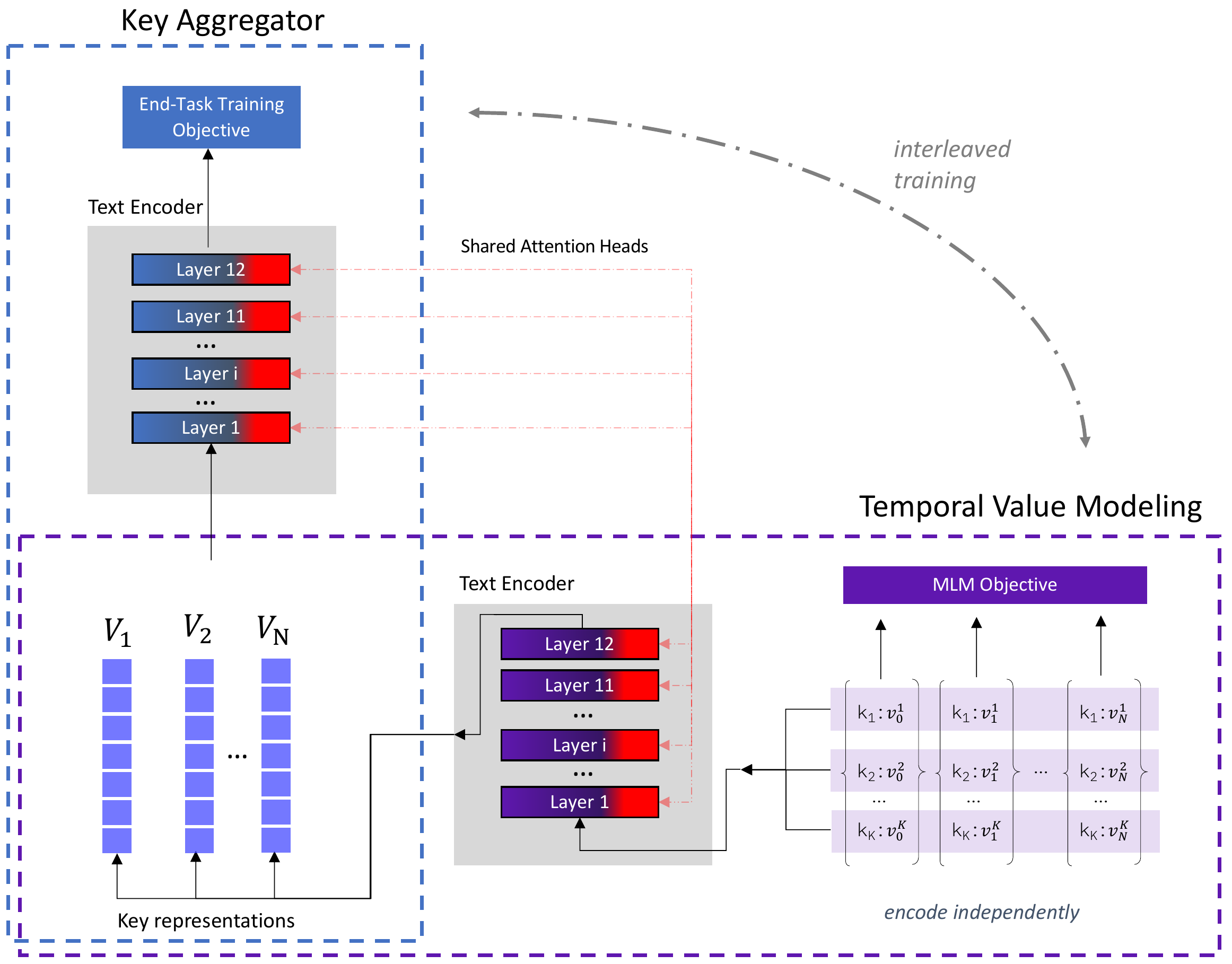}
    \caption{The TVM-KA network architecture consisting of a set of shared attention heads (weights) between the Temporal Value Modeler and the Key Aggregator. Each key is encoded independently to create a corresponding key representation.}
    \label{fig:sh}
\end{figure*}

 Let a sequence of semi-structured objects be denoted as $\mathcal{J}=[J_1, J_2, J_3, \ldots, J_N]$ 
 corresponding to $N$ time steps. Further, let $J_i = \{k_1 \colon v_{i}^1,  k_2\colon v_i^2, \ldots, k_j\colon v_i^j, \ldots, k_K \colon v_i^K\}$ denote a structured object $J_i$, containing $K$ key-value pairs $ <k_j \colon v_i^j>, j=1 \dots K$. The goal of our modeling is to learn a representation of  a sequence of structured objects $\mathcal{J}$; and subsequently, learn  $f:\text{Embd}(\mathcal{J})\rightarrow\{1,2,\ldots,C\}$ for an end task, such as a $C$-way classification task. 
 
 
 We develop a modular two-part modeling strategy to represent a sequence of structured objects. 
 \begin{enumerate}
     \item The first module, called the Temporal Value Modeler (TVM), is used to learn a combined representation (referred to as the {\em key-representations}) for the different values that each key takes in the sequence.  
     \item The second module, called the Key-Aggregator (KA), uses the {\em key-representations} corresponding to each key, to create an overall representation for $\mathcal{J}$. 
 \end{enumerate}

 \noindent \textbf{Temporal Value Modeling}: Let $k$ be a key from the universe of all the keys $\mathcal{K}$ in the sequence. Then, for each key $k$ we encode the value-aware key-representation $V_k$, by considering the value of the key $k$ at each timestamp, as a sequence. Formally, $V_k$ is given by:
 \begin{equation}
 \small
 V_k = \verb|[CLS]|  \ k \verb|[VAL_SEP]| v_1^k \ \verb|[VAL_SEP]| \ v_2^k\ \verb|[VAL_SEP]| \ldots v_N^k
\end{equation} \label{eq:val-seq}
where \verb|[VAL_SEP]| and \verb|[CLS]| are special tokens. Note that each value $v_j^k$ for a key $k$ at time step $j$ can itself consist of many tokens and those have not been shown for ease of presentation.  
With any choice of a transformer-based ~\cite{Transformer} language encoder (\textit{TextEncoder -- TE}), an embedding for $V_k$, termed the \textit{key-representation} (KR), can be obtained as:
 \begin{eqnarray}
     \text{KR}_k = \text{\textit{TextEncoder}}(V_k)[0] \label{eq:key-rep}
 \end{eqnarray}
A $TextEncoder$ gives us a $dim$ x $L$ dimension tensor where $dim$ is the output embedding size and $L$ corresponds to the length of the tokenized sequence $V_k$. We use the output embedding representation at the first position (indicated as $[0]$ in Eq. \ref{eq:key-rep}) as the {\em key-representation}. 
It is easy to see that this formulation allows us to accommodate the modeling of natural language text as in Figure \ref{fig:samples} (c). For illustration, if the $TextEncoder$ is based on BERT ~\cite{BERT}, Eq. \ref{eq:key-rep} reduces to the encoding scheme typically employed in BERT for text paragraphs, where the \verb|[VAL_SEP]| corresponds to the \verb|[SEP]| token.



 \paragraph{Key-Aggregation:}  Once we create {\em key-representations} we utilize them for an end-task. 
We encode the key-representations using the same model architecture as the TVM $TextEncoder$ but do not use positional embeddings since we encode a {\em set} of position-invariant key representations. Note that the weights of the KA are randomly initialized.
This network is directly optimized for an end-task.
 \begin{eqnarray}
    \text{Embd}(\mathcal{J}) = \text{KeyAggregator}\left(\{\text{KR}(k)\ |\  k \in \mathcal{K}\}\right) \label{eq:json}
 \end{eqnarray}

\paragraph{Key-centric vs. Record-centric Representation:} As an alternative to the {\em key-centric} representation used by the TVM, one could  construct a {\em record-centric} view to model the sequence \cite{de2021transformers4rec}. Instead of modeling the evolution of {\em keys} in a semi-structured object sequence using $V_k$ for each key, one could treat the sequence as a series of $J_i$ (Figure \ref{fig:samples}(f)). However, the record-centric representation requires the network to compress information present in multiple keys and values of a record ($J_i$) which can create an information bottleneck for the downstream task.  We compare and contrast the benefit of these alternative views in Section \ref{sec:experiments}
 and Section \ref{sec:discussion}. 

\paragraph{Challenges of Scalable Training:} The training of the hierarchical two-part network, %
which first obtains the {\em key-representations} (Eq. \ref{eq:key-rep}) and then the structured object sequence representation (Eq. \ref{eq:json}), could be done \textit{end-to-end} where the network parameters are 
directly trained for the downstream task.
However, end-to-end training of the hierarchical two-part network is often difficult due to the constraints imposed by the limited GPU memory. The GPU memory usage is affected by two factors:
(1) the length of the semi-structured object sequence; and (2) the number of keys in an object. 
The end-to-end model architecture operating over a batch of $\mathcal{J}$ sequences would exceed the memory of most commodity GPUs. By a conservative estimate, even for $n=11$ and $N=512$, a typical 120M parameter model would exceed 40GB RAM limit with a batch-size of $2$. To address this, we use an iterative training paradigm, described below, which interleaves the training of the TVM and KA components by relying on attention heads that are shared between the two components.

\paragraph{Sharing Attention Heads:} Recall that the TVM network first creates a representation for each key by attending on the values that occur in the sequence for each of them. The KA network then uses these representations to learn the end task. However, if the KA network could influence {\em how} these representations are created for each key, it could perhaps help improve the performance of the KA on the downstream-task. We, therefore, introduce hard-parameter sharing between the TVM and KA components. 
We hypothesize that by sharing a few attention heads (weights) between the two networks, the KA will be able to utilize the shared attention heads. Specifically, as training progresses and updates the parameters used in these heads, it will have an effect of adjusting the {key-representations} from the TVM in a way that could help improve overall end-task performance. 

Formally, let the TVM and KA networks use $h$ heads in multi-head attention of the transformer-based network. Let $p$ be the number of shared heads, then at any given layer in the network, the multi-head attention layers of TVM and KA are defined as:



\begin{align}
    \text{TE\_MultiHead}(Q,K,V) = {}& \text{Concat}({\color{black}\text{head}_{e_1},\dots, \text{head}_{e_p}}, \nonumber \\
    & \text{head}_{e_{p+1}},\ldots,\text{head}_{e_h})W_e^O \nonumber \\
    \text{KA\_MultiHead}(Q,K,V) = {}& \text{Concat}({\color{black}\text{head}_{a_1},\dots, \text{head}_{a_p}}, \nonumber \\ 
    & \text{head}_{a_{p+1}},\ldots,\text{head}_{a_h})W_a^O \nonumber 
\end{align}
where,
\begin{align}
    \text{head}_i = \softmax\left(\frac{QW_i^Q\left(KW_i^K\right)^T}{\sqrt{d}}\right)VW_i^V \nonumber \\
    \text{and}\ W^Q_{e_m}=W^Q_{a_m},W^K_{e_m}=W^K_{a_m}, W^V_{e_m}=W^V_{a_m}, 1\leq m\leq p \nonumber
\end{align}

Here, $W_i^Q$, $W_i^K$, $W_i^V$ are projection matrices for query, key, and value for the $i^{th}$ attention head, 
and $d$ denotes the dimension of the query and key vectors. $e_1,\ldots,e_p$ and $a_1,\ldots,a_p$ denote the $p$ pairs of attention heads that are sharing weights between the TVM and KA networks. $e_{p+1},\ldots,e_h$ denote $TextEncoder$ specific attention heads in the TVM, and $a_{p+1},\ldots,a_h$ denote KA specific attention heads. $W_e^O$ and $W_a^O$ are the output projection matrices for a layer of $TextEncoder$ and the KA.   Figure \ref{fig:sh} summarizes our parameter-sharing approach.


\paragraph{Interleaved Task Training:} 
As mentioned above, we use an iterative task training paradigm where we interleave the training of the TVM and KA components. Note that our training paradigm is different from traditional training schedules for sequential task training where one network is fully trained before the next module, or from fine-tuning approaches where a part of the network may be initialized with a pre-trained model and additional layers of the network are initialized randomly and then updated for an end-task. 
We use the Masked language modeling (MLM) objective \cite{BERT} to train the TVM component and an end-task-specific objective for training the KA. The use of interleaved training, as outlined in Algorithm \ref{algo}, prevents the problem of catastrophic forgetting \cite{CatastrophicForget1,CatastrophicForget2,CatastrophicForget3,CatastrophicForget4,CatastrophicForget5} when the KA is trained. Further, it is possible that when the TVM is trained for the first time it may rely heavily on the heads that are shared. Thus, any change to the representation from these heads could lead to poorer {\em key-representations} and attention sharing, and would therefore would be counter-productive. To address this problem, we apply DropHead ~\cite{zhou2020scheduledDrophead} on the shared attention heads in TVM and pre-train the model before beginning the interleaving schedule. 


\begin{algorithm}
\footnotesize \hspace*{\algorithmicindent} 
\caption{Interleaved training\label{algo}}

\begin{algorithmic}[1]
    \STATE \textbf{Initialize} Temporal Value Modeler  $\mathcal{M}_v$, Key Aggregator $\mathcal{M}_k$ parameters randomly. 
    
    \STATE Prepare the dataset $D_v$ consisting of value sequences 
    \FOR{i = 1,2,\dots,p}
    \STATE $\triangleright$ TVM training
    \STATE Update TVM $\mathcal{M}_v$ model parameters with MLM objective on $D_v$.
    \STATE Prepare the dataset $D_k$ consisting of key-representations $\text{KR}_k$ as per Eq. \ref{eq:key-rep}
    \STATE $\triangleright$ KA training
    \STATE Update Key Aggregator $\mathcal{M}_k$ model parameters with cross-entropy loss for downstream task.
    \ENDFOR

\end{algorithmic}
\end{algorithm}

\section{Experiments} \label{sec:experiments}

Our experiments are designed to answer the following questions: (i) How helpful is the TVM-KA architecture over the baseline that involves flattening semi-structured object sequences? (ii)  How does the model compare to existing approaches based on record-centric representations? (iii) How important is the use of shared attention heads for fine-tuning of the Key Aggregator? (iv) Does the interleaved training procedure help train the network effectively?  

\subsection{Data}
We experiment using two application/cloud logs datasets and one e-commerce purchase history dataset. The first application logs dataset, referred to as the `Cloud Service Logs,' is an internal dataset consisting of interaction traces typically used for  product usage analysis. We also use the publicly available LogHub \cite{he2020loghub} dataset, comprising system log messages from the Hadoop distributed file system, and a publicly available e-commerce dataset, which consists of product purchase information \cite{instacart}. 

\begin{table*}[!htb]
    \centering
    \resizebox{\linewidth}{!}{
    \begin{tabular}{lrrrrcrrr}
    \toprule
         \textbf{Dataset} &  \textbf{Train} & \textbf{Dev} & \textbf{Test} & \textbf{\# Classes} & \textbf{Task} & \multicolumn{1}{c}{\textbf{\# Keys}} & \multicolumn{2}{c}{\textbf{\# Time Steps}} \\
         \cmidrule(lr){8-9}
         & & & & & & & \multicolumn{1}{c}{\textbf{Median}} & \multicolumn{1}{c}{\textbf{Maximum}} \\ 
         \midrule
         Cloud Service Logs & 12,833 & 1,605 & 1,604 & 3 & Milestone Prediction & 11 & 112 (17061) & 300 (177340) \\ 
         \midrule
         LogHub & 402,542 & 57,506 & 115,012 & 2 & Anomaly Detection & 46 & 19 (1176) & 298 (18530) \\
         \midrule
         Instacart & 780,003 & 97,501 & 97,500 & 3,212 & Next Product Prediction & 10 & 134 (9842) & 3598 (267025) \\
         \midrule
        \bottomrule
    \end{tabular}
    }
    \caption{Dataset Statistics including the median and maximum length of sequences reported in number of time-steps. Values in parentheses report the sequence length after sub-word tokenization using the BERT tokenizer.}
    \label{tab:data}
\end{table*}

\paragraph{Cloud Service Logs Data -- Application event traces from a large cloud provider:}
In the Cloud Service Logs dataset, application event traces are logged by the cloud provider website. 
Event types include login, browsing, account creation/maintenance/update, UI navigation, search, service creation/deletion, app interactions, and others. 
Each event has an associated payload that provides context around the event. 
Our raw data is a snapshot of application event traces spanning 3 months and comprising about 450M events, from which we build our user sessions. A user session is essentially a temporal sequence of event traces for that user. While the raw data has over $60$ keys in each event, we use a smaller set of manually selected $11$ keys, so that existing approaches and baselines can be meaningfully used for comparison.     
We constructed user sessions for 100k users. The application events corresponding to 1) plan upgrade, and 2) opening chatbot (to seek help) are considered as \textit{milestone events}. These milestone events are chosen to represent revenue generation and user experience, respectively. 
The case of no milestone event occurring is treated as third class. 
From the traces, temporal sequences of 300 events are extracted to predict if a milestone (or no-milestone) event will occur in next 50 time steps.  We report Macro F1-Score, as the dataset exhibits class imbalance. We include additional details about the dataset in the appendix.

\paragraph{Instacart eCommerce Data:}
The publicly available Instacart dataset\footnote{https://tech.instacart.com/3-million-instacart-orders-open-sourced-d40d29ead6f2} contains $3$ million grocery purchase orders of nearly $200,000$ users of the application. Each order consists of multiple products and each structured object associated with a product contains meta-data such as day of the week, product category, department, aisle, etc. We reprocess this dataset to create sequences of product purchases and evaluate models on the task of the next product prediction. 
We predict the product name given the sequence of product orders,\footnote{We use the complete structured object.} which is effectively a classification task over a universe of $3212$ products. Existing work on this dataset has focused on a simpler binary prediction task where models are asked to predict if a particular item is likely to be purchased again.\footnote{https://www.kaggle.com/competitions/instacart-market-basket-analysis/leaderboard}

\paragraph{LogHub Data:}
We use the HDFS-1 from LogHub \cite{he2020loghub} for the log anomaly detection task. As the dataset originally consisted of lines of log messages, we use the Drain log parser \cite{he2017drain} to identify $48$ log templates. Using a semi-automated approach, we assign key names to the value slots of the templates. Thus, each log line is converted to a structured object with $46$ key-value pairs. The original dataset splits the log lines into \textit{blocks,} and the binary prediction task is to predict whether a particular block is {\em anomalous}. The dataset is highly imbalanced, with around $3\%$ of the instances belonging to the anomalous class. So, we report F1-Score for the anomalous class.

\subsection{Encoders}
\noindent {\bf Baselines:} We flatten each key-value pair in a structured object and encode them with special markers indicating boundaries for objects and timesteps. We fine-tune the pre-trained encoders for each downstream task and report their performance. We experiment with BERT \cite {BERT} and Longformer \cite{Longformer} as the pre-trained encoders. 

We also compare our model with popular approaches for creating record-centric representations. 
These approaches first obtain the representations for each object $J_i$, and then feed them to an inter-object transformer to create the representation of the whole sequence of objects.
We experiment with three popular methods,
where each object representation is obtained from the key-value pair representations by 1) point-wise summation, 2) concatenation then project-down~\cite{Mizrachi2019CombiningCF,de2021transformers4rec}, and 3) self-attention then averaging~\cite{zhang-etal-2019-hibert,gu2021dialogbert}.


\paragraph{Encoders for TVM-KA:} One of the advantages of the TVM-KA architecture is that it is agnostic to the choice of the encoder. We employ the same encoder architectures used in our baselines to enable a direct performance comparison. Recall that the TVM module and KA module share attention heads to facilitate sharing of information between them. 
To pre-train the TVM, we mask 15\% of the tokens in every Value Sequence, and the objective is to predict the masked tokens. We do not mask \textit{value-separator} and \textit{key aggregator} tokens; we only mask the values. Details on hyper-parameter tuning and the iterative training schedule are available in the appendix. 




\subsection{Results}
Table \ref{tab:all_results} reports the primary results from our experiments. We include the results on three datasets and for each dataset, we report the overall performance along with the performance of the models on slices of the dataset where the length of the sequence is greater than the median length for that dataset.

\begin{table*}[!htb]
\centering
\resizebox{\textwidth}{!}{%
\begin{tabular}{@{}llcccccc@{}}
\toprule
\multirow{2}{*}{} & \multicolumn{1}{c}{\multirow{2}{*}{\textbf{Configuration}}} & \multicolumn{2}{c}{\textbf{Cloud Service Logs (Macro F1)}} & \multicolumn{2}{c}{\textbf{Instacart (Recall@10)}} & \multicolumn{2}{c}{\textbf{Loghub (Binary F1-Score)}} \\ \cmidrule(lr){3-4} \cmidrule(lr){5-6} \cmidrule(lr){7-8} 
 & \multicolumn{1}{c}{} & \begin{tabular}[c]{@{}c@{}}{\textbf{L \textgreater Median}} \\($50\%$)\end{tabular} & \multicolumn{1}{c}{\textbf{Overall}} & \begin{tabular}[c]{@{}c@{}}{\textbf{L \textgreater Median}} \\($50\%$)\end{tabular} & \multicolumn{1}{c}{\textbf{Overall}} & \begin{tabular}[c]{@{}c@{}}{\textbf{L \textgreater Median}} \\($51.37\%$)\end{tabular} & \multicolumn{1}{c}{\textbf{Overall}} \\ \midrule
\multirow{3}{*}{\begin{tabular}[c]{@{}l@{}}Flattened Encoding \\ (BERT) \\ \cite{BERT} \end{tabular}} & \begin{tabular}[c]{@{}l@{}}Random\\ (bert-base-uncased)\end{tabular} & 49.55 & 50.22 & 9.6 & 9.4 & 0.00 & 53.62 \\
\cmidrule{2-8}
 & \begin{tabular}[c]{@{}l@{}}Pre-Trained\\ (bert-base-uncased)\end{tabular} & 77.30 & 74.77 & 20.10 &  18.70 & 23.32 & 61.63 \\
 \cmidrule{2-8}
 & \begin{tabular}[c]{@{}l@{}}Pre-Trained\\ (bert-large-uncased)\end{tabular} & 80.06 & 76.59 & 21.07 &  19.11 & 58.30 & 75.86\\
 \midrule
\begin{tabular}[c]{@{}l@{}}Flattened Encoding \\ (Longformer) \\ \cite{Longformer}\end{tabular} & Pre-Trained & 75.83 & 73.71 & 16.94 & 16.08  & 97.71 & 98.54 \\
 \midrule
\begin{tabular}[c]{@{}l@{}}Flattened Encoding \\ (GPT-2) \end{tabular} & Pre-Trained & - & 64.92 & - & - & - & - \\
\midrule
\midrule
\multirow{3}{*}{\begin{tabular}[c]{@{}l@{}}Record-centric\\ Representation \\ \cite{de2021transformers4rec}\\ \cite{gu2021dialogbert} \end{tabular}} & Summation & 78.97 & 77.33 & 7.11 & 5.10 & 99.22 & 99.51 \\
\cmidrule{2-8}
 & Concat & 77.76 & 75.99 & 7.18 & 5.11 & \textbf{99.29} & \textbf{99.57} \\
 \cmidrule{2-8}
 & Self-Attention & 79.18 & 77.73 & 7.98 & 6.34 & 99.08 & 99.42 \\
 \midrule
\midrule
\multirow{3}{*}{TVM-KA} & Joint Modeling & 80.15 & 77.68 & 17.04 & 16.0 & 46.78 & 70.72 \\
\cmidrule{2-8} 
& No Interleaving & 73.19 & 73.22 & 18.32 & 17.5 & 99.05 &  98.64 \\
\cmidrule{2-8} 
 & Interleaving & \textbf{81.26} & \textbf{79.60} & \textbf{23.44} &  \textbf{22.54} & 98.79 &  99.32 \\ 
 \bottomrule
\end{tabular}%
}
\caption{Comparison of TVM-KA model with the baseline approaches on various datasets. In our proposed approach, both TVM and KA components have the same architecture and the number of parameters as \textit{bert-base-uncased}. We additionally report results on a subset of the test set whose sequence length (L) (post-tokenization) is greater than the median length for each dataset. The values in parenthesis indicate the percentage number of instances where the length of a sequence is greater than the median sequence length. The results from our approach are statistically significant with respect to all other approaches on both cloud service logs and instacart datasets (p-value $<$ 0.03). }
\label{tab:all_results}
\end{table*}

\paragraph{Comparison with Flattened encoding:} 
As seen in the `overall' scores for each dataset in Table \ref{tab:all_results}, flattening (first four rows) yields a significantly lower performance compared to our approach involving the use of interleaved training for TVM-KA (last row). For instance, on the cloud service logs, there's an increase of 3.9\%-58.5\% compared to flattened encodings in macro F1 scores. Similar trends are reported on the Instacart dataset. 

Interestingly, we find that on the LogHub dataset, the Longformer model \cite{Longformer} is able to obtain a comparable performance (98.54 vs 99.32). A deeper investigation reveals that the Loghub dataset, unlike our other datasets, has a median maximum sequence length (post sub-word tokenization) of $1176$ tokens. Since the Longformer model can handle sequences of up to $4096$ tokens, it is capable to perform on par with our TVM-KA approach.



\paragraph{Comparison with Record-centric representations:} 
Unlike flattened encoding, the record-centric representation does not suffer from the modeling limitations associated with the maximum sequence length limit. These representations can encode sequences in their entirety, since most datasets have fewer than 300 objects (time-steps), and the sequence length is equal to the number of time steps. However, the record-centric view may not adequately model the dependencies between values of the same key across different time steps. We find that our approach outperforms all record-centric representation baselines on the Cloud Service Logs dataset as well as the Instacart dataset. Interestingly, the performance of TVM-KA on the LogHub data is similar to that of different record-centric baselines (99.32 vs 99.57) which may be due to the relative simplicity of the prediction task. 

\begin{figure}[!htb]
    \centering
    \includegraphics[width=\columnwidth]{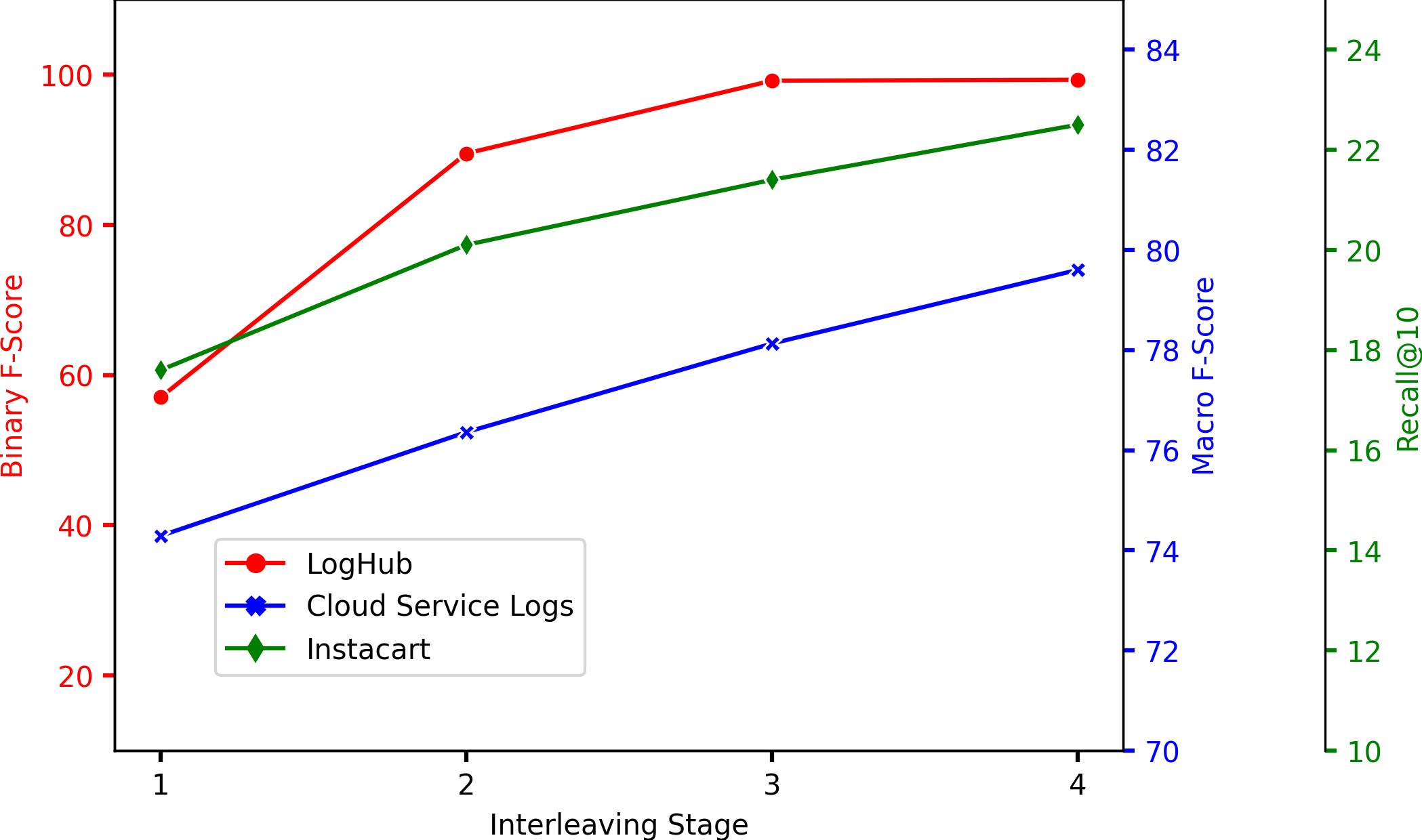}
    \caption{TVM-KA model performance for each interleaving stage. The red, green, and blue lines, along with their respective colored y-axes, indicate the performance of the Loghub, Instacart, and Cloud Service Logs datasets, respectively.}
    \label{fig:tvmkaTime}
\end{figure}
\paragraph{Importance of interleaved training:}
As seen in Table \ref{tab:all_results},
using our interleaving training method outperforms training the model with no interleaving.
This supports our hypothesis that sharing a few attention heads helps the TVM-KA model uncover better key-representations for the downstream task as training progresses. Figure \ref{fig:tvmkaTime} illustrates the increase in performance with each stage of interleaving on all three datasets.


\paragraph{Joint Modeling vs Interleaved Training of TVM-KA:} We observe that \textit{joint modeling} performs poorly compared to our interleaving approach. 
We found this surprising as we had expected it to be at par with our approach when the joint models fit in memory.\footnote{We trained our joint models on 80GB A100 GPUs.} We hypothesize that by interleaving and sharing attention heads between TVM and KA, the fine-tuning of the KA on the downstream task introduces a task-specific bias to help improve the key-representations. This in-turn, benefits the pre-trained representations of the TVM via shared-attention weights and further improves performance in the next round of training. In the absence of this bias, the joint model perhaps converges at an alternative minima that is not as good. 




\paragraph{Effect of varying the number of shared attention heads:} In general, we observe that sharing of $4$ and $6$ attention heads helps the most. Sharing too few or too many attention heads results in an average drop of 2\%-43\% in performance. We include further details in the appendix. 

\paragraph{ Effect of parameter size/model capacity:} To investigate if the model capacity could be a bottleneck for the approaches based on flattened representations. We fine-tune \textit{bert-large-uncased} model which has $3x$ the parameter of \textit{bert-base-uncased} model and approximately $1.5x$ the parameters of the TVM-KA network. We find that our model performs better and hypothesize that the improved performance is primarily due to the value-aware key representations and the model's resultant ability to accommodate longer sequences due to the decoupling of keys. 

\section{Related Work}
Our work is related to several areas of research.
First, modeling multiple sets of key-value entries has similarities to modeling the rows and columns in a table. Second, the predictive tasks we apply our models to are related to multi-variate regression and spatio-temporal modeling tasks. Finally, our two-stage TVM-KA architecture is a type of hierarchical encoder, forms of which have been used for multiple tasks, from modeling tabular data to encoding text. 

\paragraph{Modeling Tabular and Timeseries Data:} 
As we explicitly model the value sequence for each key, the data object we work with is reminiscent of tabular data where each row is a time-step and each column comprises the values for a particular field.
On the surface, the modeling of data may appear related, but the actual tasks and models developed for tasks on tabular data cannot be applied to semi-structured object sequences. This is because the work on modeling textual tabular data often involves developing specialized models focused on retrieving information from cells ~\cite{Zayats2021RepresentationsFQ,wang2021retrieving,iida2021tabbie,yang2022tableformer}, multi-hop reasoning across information in different cells across parts of the table ~\cite{chen-etal-2021-finqa,zhao2022multihiertt}, combining information present in tables and unstructured text for information seeking tasks~\cite{HanboLi2021,zhu2021tat,Zayats2021RepresentationsFQ,luetto2023one,cholakov2022gatedtabtransformer}, etc. In addition, work on modeling time-series tabular data has focused on numerical data \cite{zhou2021informer,zerveas2021transformer,zhao-etal-2022-multihiertt} with purpose-built task-specific architectures that cannot be easily adapted to other tasks~\cite{wu2021autoformer,padhi2021tabular}. 

\paragraph{Modeling Temporal Graph Sequences and Recommender Systems:} Our approach is also related to a rich body of work on modeling temporal graphs and recommendation systems~\cite{xu2021anomaly,grigsby2021long,de2021transformers4rec}. Temporal graph evolution problems involve constructing representations to enable tasks such as link prediction \cite{dysat,Xu_Liang_Cheng_Wei_Chen_Zhang_2021}, item recommendation in user sessions \cite{Hsu2021RetaGNNRT}, answering queries on graphs and sequences \cite{saxena2022sequence}, classifying graph instances in a sequence,\cite{Xu_Liang_Cheng_Wei_Chen_Zhang_2021,xu2021anomaly} etc. Our findings suggest that such approaches do not scale for long sequences for the tasks we experimented with. However, our record-centric model baselines \cite{de2021transformers4rec,Mizrachi2019CombiningCF} are similar in approach to these methods. 
\paragraph{Parameter sharing in neural network models:} Deep neural networks are usually trained to tackle different tasks in isolation. Networks that cater to multiple related tasks (multi-task neural networks) seek to improve generalization and process data efficiently through parameter sharing and joint learning. Traditional hard-parameter sharing uses the same initial layers and splits the network into task-specific branches at an ad hoc point~\cite{guo2018dynamic,lu2017fully}. On the other hand, soft-parameter sharing shares features via a set of task-specific networks~\cite{liu2019end,maninis2019attentive}. More recently adaptive sharing approaches have been proposed that decide what parameters to share across tasks to achieve the best performance~\cite{vandenhende2019branched,sun2020adashare}. The parameter sharing utilized in this work is different from the aforementioned approaches, as we share some attention head weights between the two networks (as compared to shared layers), in a way that could help to improve overall end-task performance.


\section{Discussion and Conclusion} \label{sec:discussion}
In this paper, we have presented a two-part encoder to model structured object sequences. 
The choice of a key-centric representation enables us to encode larger objects as well as long sequences. Our experiments show that by using the two-part TVM-KA architecture, we are able to inject downstream task information into the temporal value modeler network to generate key representations that are more relevant. 
However, the key-centric representation does not allow the model to support tasks such as sequence tagging of the structured objects. Nor does it allow to model graph sequences effectively, as it  does not use a global view of the structured objects. Thus, it may not be able learn patterns {\em across fields} at different time steps. For such tasks, a record-centric representation is perhaps more helpful. Both key-centric and record-centric representations have their strengths and weaknesses, and the choice should be made with the downstream task in mind.  

We additionally present a novel interleaving scheme to train our two-part encoder. We induce task bias into the model by sharing attention heads between Temporal Value Modeler and Key Aggregator components. Our proposed approach outperforms the baseline approaches which flatten structured object sequences and those based on record-centric representations. To the best of our knowledge, we are the first to demonstrate the use of a key-centric representation for
structured object sequence encoding at scale. 
\bibliography{emnlp2023}
\bibliographystyle{acl_natbib}

\appendix


\section{Dataset Details}
We provide additional details about the datasets used in our experiments.

\subsubsection{Cloud Service Logs Data: Application event traces from a large cloud provider}
In the Cloud Service Logs (CSL) dataset, application event traces are logged in the cloud provider website. Each user is assigned a unique identifier. Event types range from \texttt{login}, \texttt{browsing}, \texttt{account creation}, \texttt{account maintenance}, \texttt{account update}, \texttt{UI navigation}, \texttt{search}, \texttt{service creation}, \texttt{service deletion}, \texttt{app interactions}, among several others. We have about $638$ unique event types. Each event has an associated payload that provides context around the event. For example, if a user performed a search, the payload captures the search query and the page where the search was performed. If a user interacted with a service, the payload captures the service ID and action performed on the service, among other information.

Our raw data is a snapshot of application event traces spanning 3 months comprising about 450M events. Using these, we build our user sessions. A user session is essentially a temporal sequence of event traces for that user. If there is a difference of greater than 15 minutes between two consecutive events, we break the session. We constructed user sessions for 100k users. The application events corresponding to 1) plan upgrade, and 2) opening chatbot (to seek help) are considered as \textit{milestone events}. These milestone events are chosen as they represent revenue generation and user experience, respectively. 
The case of no milestone event occurring is treated as the third class.

From the traces, we identify user sessions containing any of the aforementioned milestone events. We consider the temporal sequences of events $350$ time-steps before the milestone event occurs. To construct the data, we consider the sequence of events till $300$ time-steps and the task is to predict if a milestone (or no milestone) event will occur in the next 50 time steps. 

\subsubsection{Instacart eCommerce Data}
The publicly available Instacart dataset\footnote{https://tech.instacart.com/3-million-instacart-orders-open-sourced-d40d29ead6f2} contains $3$ million grocery purchase orders of nearly $200,000$ users of the application. Each order consists of multiple products and each structured object associated with a product contains meta-data such as the day of the week, the product category, department, aisle, etc. We reprocess this dataset to create sequences of product purchases and evaluate models on the task of the next product prediction. We create variable-length training instances from each user's order history by sampling between 50 to 200 previous product purchases for a certain target product. Additionally, we only sample a training instance if the target product has been ordered at least 50 times across users.  

As per our task formulation, we predict the product name given the sequence of product orders\footnote{We use the complete structured object.}, which is effectively a classification task over a universe of $3212$ products. Existing work on this dataset has focused on a simpler binary prediction task where models are asked to predict whether a particular item is likely to be purchased again.\footnote{https://www.kaggle.com/competitions/instacart-market-basket-analysis/leaderboard}


\subsubsection{LogHub Data}
HDFS-1 from LogHub \cite{he2020loghub} is utilized for the log anomaly detection task. The dataset consists of log lines. A sample log message is shown below:
\begin{lstlisting}[breaklines]
081109 203519 29 INFO dfs.FSNamesystem: BLOCK* NameSystem.addStoredBlock: blockMap updated: 10.250.10.6:50010 is added to blk_-1608999687919862906 size 91178
\end{lstlisting}
We now convert the log message to a JSON object. We utilize Drain log parser \cite{he2017drain} to extract the static template, dynamic variables, and header information from log messages. We obtain around $48$ templates. All the log messages fall into one of the $48$ templates. In a semi-automated fashion, we define keys for the templates. We populate key names with the value slots of the templates. Thus, each log line is converted to a structured object. The log message after conversion to a JSON object would look as follows,

\begin{json}
\begin{lstlisting}[language=json]
{
    "status": "addStoredBLock: 
    Blockmap updated",
    "port": "10.250.10.6:50010",
    "block_ID": "blk_-1608999687919862906",
    "size": "91178",
    "LineId": "11",
    "Date": "81109",
    "Time": "203519",
    "Pid": "29",
    "Level": "INFO",
    "Component": "dfs.FSNamesystem",
    "EventId": "5d5de21c"
}
\end{lstlisting}
\end{json}

A \textit{block} consists of a sequence of such structured objects. The task is to classify the given block as anomalous or not.

\section{Hyperparameters and Training schedule}
We perform a grid search for the learning rate and batch size for all the models in our experiments. We select the hyper-parameter configuration which gives the best validation set performance on each dataset's metric. We now list the range of values considered for each hyper-parameter.

\begin{itemize}
    \item \textbf{Learning Rate:} $\{1e^{-4}, 3e^{-4}, 5e^{-4}, 1e^{-5}, 3e^{-5}, \\
    5e^{-5}, 1e^{-6}, 3e^{-6}, 5e^{-6} \}$,
    \item \textbf{Batch Size:} $2, 4, 8, 16, 32$
    \item \textbf{Shared Heads (p):} $2, 4, 6, 8$
\end{itemize}

The drophead mechanism is only activated during TVM training, with drophead probability set to $0.2$.

For all the baseline models, we train till convergence. For the TVM training in the first iteration, we vary learning rates and observe model convergence (in terms of train and dev loss) after a fixed $100K$ steps. The best learning rate for TVM is identified from this exercise. A similar approach is used for identifying the best learning rate for the KA training stage too, albeit for a smaller number of update steps. In the first iteration, TVM training is done for $1$ epoch. Then we proceed with the interleaving step. We alternate between TVM training and KA training with their number of training steps in 2:1 proportion. For the cloud service logs dataset, the number of TVM training steps is 50K and the number of KA training steps is around 100K.

\section{Class-wise Results on Cloud Service Logs}
Table \ref{tab:cloud-service-classwise} reports the detailed class-wise results on the Cloud Service Logs dataset. As seen in the `Macro F1' score column in the Table \ref{tab:cloud-service-classwise}, flattening (first four rows) yields a significantly lower performance compared to our approach involving the use of interleaved training for TVM-KA (last row). Specifically, the flattened encoding with randomly initialized BERT model suffers the most on the \texttt{Open Chatbot} class. We believe that semantic understanding of the event names is crucial for identifying the sequences leading to \texttt{Open Chatbot} milestone event.

In general, we observe improvements from our TVM-KA approach on all class labels compared to the baseline models. This indicates the performance gain from our model is not due to improvements in a single class or subset of classes, but, on all the classes present in our dataset.

\section{Influence of shared attention heads} 
We use BERT encoder \cite{BERT} to model both TVM and KA components in our model. This allows us to share attention heads between the TVM and KA components. \textit{bert-base-uncased} has $12$ attention heads at each encoder layer. We experiment with sharing $0, 2, 4, 6, 8$ attention heads between TVM and KA components.

\begin{figure}[!htb]
    \centering
\includegraphics[width=0.45\textwidth]{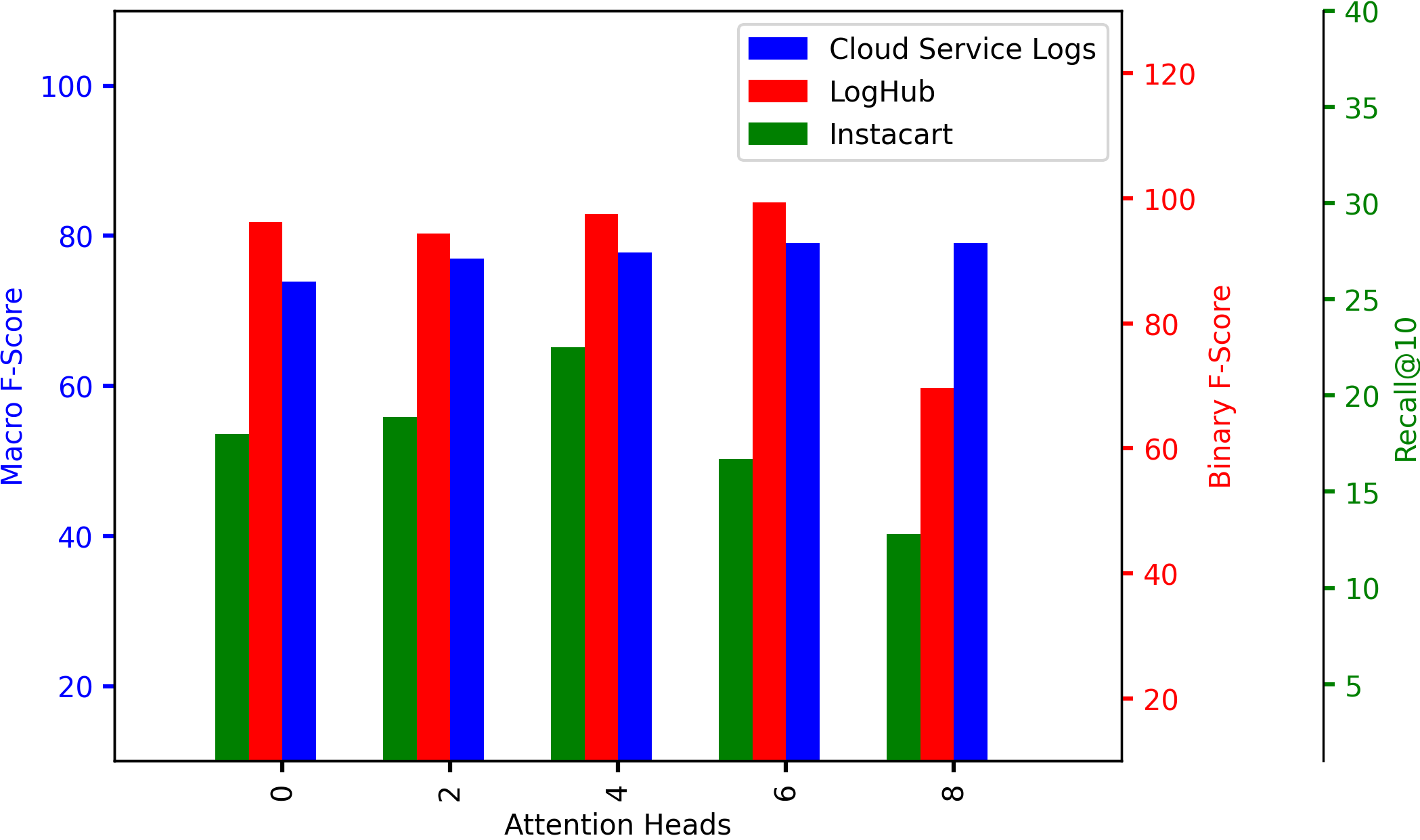}
    \caption{Effect of sharing different attention heads between TVM and KA}
    \label{fig:shared_heads}
\end{figure}
Figure \ref{fig:shared_heads} presents the performance of our TVM-KA approach with different numbers of shared attention heads between TVM and KA components. In general, we observe that sharing of $4$ and $6$ attention heads helps the most. While not sharing any attention heads or sharing more results in poor performance.

\begin{table*}[!htb]
\resizebox{\linewidth}{!}{%
\begin{tabular}{@{}llrrrrr@{}}
\toprule
 & \multicolumn{1}{c}{\textbf{Comments}} & \multicolumn{1}{c}{\textbf{Macro F1}} & \multicolumn{1}{c}{\textbf{Micro F1}} & \multicolumn{1}{c}{\textbf{\begin{tabular}[c]{@{}c@{}}Browsing/\\ Upgrade Account\end{tabular}}} & \multicolumn{1}{c}{\textbf{No MileStone}} & \multicolumn{1}{c}{\textbf{Open Chatbot}} \\ \midrule
\multirow{3}{*}{\begin{tabular}[c]{@{}l@{}}Flattened Encoding \\ (BERT) \\ \cite{BERT} \end{tabular}} & \begin{tabular}[c]{@{}l@{}}Random \\ (bert-base-uncased) \end{tabular}  & 50.22 & 72.38 & 70.14 & 80.50 & 0.0  \\
\cmidrule{2-7}
& \begin{tabular}[c]{@{}l@{}}Pre-Trained \\ (bert-base-uncased) \end{tabular}  & 74.77 & 80.31 & 79.39 & 84.12 & 60.79 \\
 \cmidrule{2-7}
 & \begin{tabular}[c]{@{}l@{}}Pre-Trained \\ (bert-large-uncased) \end{tabular} & 76.59 & 81.57 & 80.72 & 85.13 & \textbf{63.93} \\
  \midrule
\begin{tabular}[c]{@{}l@{}}Flattened Encoding \\ (Longformer) \\ \cite{Longformer}\end{tabular} & Pre-Trained  & 73.71 & 79.91 & 79.31 & 84.26 & 57.57 \\
 \midrule
 \multirow{3}{*}{\begin{tabular}[c]{@{}l@{}}Record-centric\\ Representation \\ \cite{de2021transformers4rec}\\ \cite{gu2021dialogbert} \end{tabular}} & Summation & 77.33 & 85.66 & 84.91 & 90.59 & 56.50 \\
 \cmidrule{2-7}
 & Concat & 75.99 & 85.12 & 84.47 & 90.15 & 53.36 \\
 \cmidrule{2-7}
 & Self-Attention & 77.73 & 85.66 & 84.78 & 90.48 & 57.92 \\
\midrule
\multirow{3}{*}{TVM-KA} & Joint Modeling & 69.61 & 80.54 & 80.04 & 86.82 & 41.97 \\
 \cmidrule{2-7}
& No Interleaving & 73.22 & 84.14 & 83.03 & 90.32 & 46.31 \\
\cmidrule{2-7}
 & \textbf{Interleaving} & \textbf{79.60} & \textbf{86.49} & \textbf{85.47} & \textbf{91.24} & 62.11 \\
 \bottomrule
\end{tabular}%
}
\caption{Comparison of TVM-KA model with the baseline approaches on Cloud Service Logs datasets. In our proposed approach, both TVM and KA components have the same architecture and the number of parameters as \textit{bert-base-uncased}. We additionally report class-wise results. The results from our approach are statistically significant with respect to all other approaches (p-value $<$ 0.03). }
\label{tab:cloud-service-classwise}
\end{table*}

\label{sec:appendix}

This is a section in the appendix.

\end{document}